\documentclass[sigconf]{acmart}
\AtBeginDocument{%
  }

\setcopyright{rightsretained}
\copyrightyear{2025}
\acmYear{2025}
\acmDOI{XXXXXXX.XXXXXXX}
\acmConference[CIKM'25]{The 33rd ACM International Conference on Information and Knowledge Management}{Nov 10--14, 2025}{Coex, Seoul, Korea}

\usepackage{graphicx}
\usepackage{lipsum}
\usepackage{xspace}
\usepackage{booktabs}
\usepackage{multicol}
\usepackage{multirow}
\usepackage{array}
\usepackage{color}
\usepackage{amsmath}
\usepackage{colortbl}
\usepackage{pifont}
\usepackage{quoting}
\quotingsetup{leftmargin=1em, rightmargin=1em}

\begin{document}

\title{\rewardmodel: An Automated Metric for Multimodal Knowledge Acquisition in Multimedia Learning}

\author{Joy Lim Jia Yin}
\orcid{0009-0006-7971-6096}
\author{Daniel Zhang-Li}
\orcid{0009-0009-3681-1896}
\author{Jifan Yu}
\orcid{0000-0003-3430-4048}
\author{Haoxuan Li}
\email{lin-jy23@mails.tsinghua.edu.cn}
\email{zlnn23@mails.tsinghua.edu.cn}
\email{yujifan@tsinghua.edu.cn}
\email{lucasli@buaa.edu.cn}
\affiliation{%
  \institution{Tsinghua University}
  \institution{Beihang University}
  \city{Beijing}
  \country{China}
}



\author{Shangqing Tu}
\author{Yuanchun Wang}
\author{Zhiyuan Liu}
\email{tsq22@mails.tsinghua.edu.cn}
\email{wangyuanchun@ruc.edu.cn}
\email{liuzy@tsinghua.edu.cn}
\affiliation{%
  \institution{Tsinghua University}
  \institution{Renmin University of China}
  \city{Beijing}
  \country{China}
}



\author{Huiqin Liu}
\orcid{0000-0002-5754-2623}
\author{Lei Hou}
\orcid{0000-0002-8907-3526}
\author{Juanzi Li}
\orcid{0000-0002-6244-0664}
\author{Bin Xu}
\authornotemark[2]
\email{liuhq@tsinghua.edu.cn}
\email{houlei@tsinghua.edu.cn}
\email{lijuanzi@tsinghua.edu.cn}
\email{xubin@tsinghua.edu.cn}
\affiliation{%
  \institution{Tsinghua University}
  \city{Beijing}
  \country{China}
}



\renewcommand{\shortauthors}{Joy Lim Jia Yin et al.}
\newcommand{\rewardmodel}{{\textit{LecEval}}\xspace}

\begin{abstract}
Evaluating the quality of slide-based multimedia instruction is challenging. Existing methods like manual assessment, reference-based metrics, and large language model evaluators face limitations in scalability, context capture, or bias. In this paper, we introduce \rewardmodel, an automated metric grounded in Mayer’s Cognitive Theory of Multimedia Learning, to evaluate multimodal knowledge acquisition in slide-based learning. \rewardmodel assesses effectiveness using four rubrics: Content Relevance (CR), Expressive Clarity (EC), Logical Structure (LS), and Audience Engagement (AE). We curate a large-scale dataset of over $2,000$ slides from $50+$ online course videos, annotated with fine-grained human ratings across these rubrics. A model trained on this dataset demonstrates superior accuracy and adaptability compared to existing metrics, bridging the gap between automated and human assessments. We release our dataset and toolkits at \url{https://github.com/JoylimJY/LecEval}.
\end{abstract}




\maketitle

\section{INTRODUCTION}

\begin{figure*}[ht]
\centering
  \includegraphics[width=0.9\linewidth]{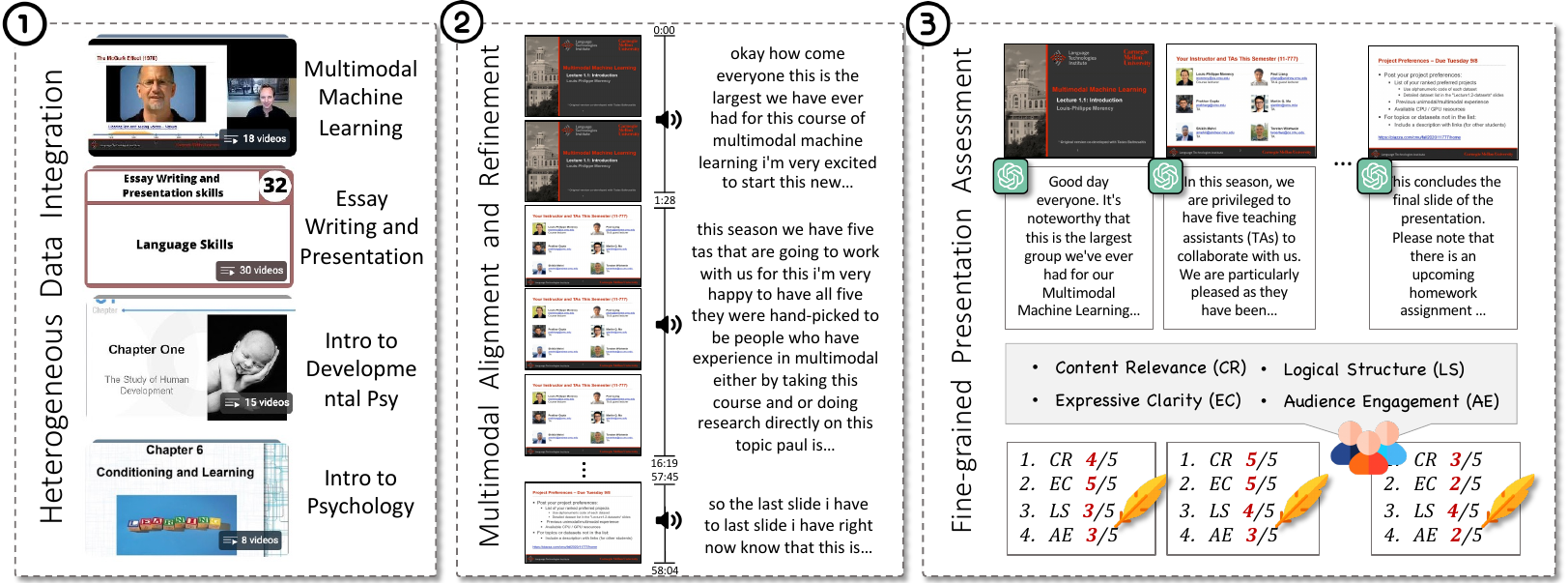}
  \caption{The general data construction framework of our dataset. (1) \textbf{Heterogeneous Data Integration}: We collect and process online lecture videos, extracting both slides and speech transcripts. (2) \textbf{Multimodal Alignment and Refinement}. We manually align transcriptions with their corresponding slides and refine the raw transcriptions leveraging GPT-4. (3) \textbf{Fine-grained Assessment}. We engage experienced human annotators to evaluate the slide presentations across our predefined rubrics.}
  \label{fig:rewardmodel}
\end{figure*}

Slide presentations are central to modern multimedia education~\cite{jenkins2009confronting,buckingham2007digital,buckingham2013media,strauss2011optimizing,brock2011empowering}, yet evaluating their learning effectiveness remains a significant challenge without standardized benchmarks~\cite{leon2021impact}. Traditional assessment methods, such as student feedback or expert reviews, are often subjective, labor-intensive, and inadequate for the dynamic, large-scale nature of e-learning environments~\cite{pereira2016assessment,yu2021mooccubex,hill2013learning,somayeh2016effectiveness}. This underscores the need for standardized, automated, data-driven metrics to reliably assess the quality of slide-based learning, including multimodal knowledge acquisition~\cite{angrist2024improve}.
While reference-based metrics like BLEU~\cite{bleu2002}, ROUGE~\cite{lin-2004-rouge}, and METEOR~\cite{banerjee2005meteor} automate text evaluation using human references~\cite{graham2015re,rouge-1,rouge-2,rouge-3}, their applicability to slide-based learning is limited~\cite{rouge-weakness-1,rouge-weakness-2}, as they depend on predefined references, impractical for dynamic content, and employ static rubrics ill-suited for context-specific educational material.
More recently, Large Language Models (LLMs) like GPT-4~\cite{openai2024gpt4technicalreport} enable reference-free evaluation~\cite{fu2023gptscore,wang2024pandalm,kim2024prometheus}, offering scalable and adaptable assessment for dynamic contexts~\cite{wang-demszky-2023-chatgpt}. Nevertheless, LLMs still struggle to replicate human nuance~\cite{wang-demszky-2023-chatgpt}, facing challenges in deep contextual understanding, interpreting custom rubrics, and aligning with human judgment~\cite{stureborg2024large,shen-etal-2023-large,chiang-lee-2023-large}.

In this paper, we propose \rewardmodel, an automated metric for evaluating multimodal knowledge acquisition in slide-based learning, inspired by Mayer's Cognitive Theory of Multimedia Learning~\cite{multimedia-mayer}. We employ four critical rubrics: \emph{Content Relevance (CR)}, \emph{Expressive Clarity (EC)}, \emph{Logical Structure (LS)}, and \emph{Audience Engagement (AE)}. We curate a dataset of over $2,000$ slide-text pairs from $50+$ online course videos, with fine-grained human annotations per rubric. A model trained on this dataset demonstrates that \rewardmodel captures human evaluation granularity and significantly outperforms existing metrics in effectiveness and reliability, offering a robust, scalable solution for evaluating multimodal educational content.

\textbf{Contributions.} Our main contributions are:
(1) We propose \rewardmodel, a novel metric for evaluating slide-based multimodal learning.
(2) We release a free, high-coverage, and fine-grained annotated dataset from online lectures with toolkits for analysis.
(3) We train a model on the dataset that significantly outperforms existing evaluation baselines.
(4) We present extensive experiments demonstrating the effectiveness and practical value of \rewardmodel.

\section{BACKGROUND}


\textbf{NLG Evaluation:} (1) \textit{Reference-based Metrics.} These evaluate generated text against human references~\cite{bernardi2016automatic,kilickaya2016}, including widely used metrics like \emph{BLEU}~\cite{bleu2002}, \emph{ROUGE}~\cite{lin-2004-rouge}, \emph{METEOR}~\cite{banerjee2005meteor}, \emph{CIDEr}~\cite{vedantam2015cider}, and \emph{SPICE}~\cite{anderson2016spice}~\cite{zhang2024rouge,hamdhana2024improved,liu2017improved,herdade2019image,wang2019describing}. Although computationally efficient, they rely on surface-level comparisons, struggling with semantic meaning~\cite{vasilyev-etal-2020-fill}. Their main drawbacks include requiring gold-standard references and poor performance with complex or domain-specific content, particularly without adequate references~\cite{rouge-weakness-1,rouge-weakness-2,yi-etal-2020-improving,fu2023gptscore,jiang2024tigerscore,zhang2024towards}.
(2) \textit{Reference-free LLM Evaluators.} Advances in LLMs like GPT-4~\cite{openai2024gpt4technicalreport} enable automated, reference-free evaluation of natural language text, including cross-modal assessment. Key approaches~\cite{gao2024llmbased} involve using pre-trained models as \emph{LLM-derived metrics}~\cite{zhang2020bertscore,fu2023gptscore,jia-etal-2023-zero,xie-etal-2023-deltascore}, employing structured \emph{LLM prompting methods}~\cite{liu2023geval,yuan2023batcheval,bai2024benchmarking,xie-etal-2024-doclens}, and developing \emph{fine-tuned LLM evaluators} trained on specific tasks~\cite{ke-etal-2024-critiquellm,wang2024pandalm,kim2024prometheus,xu-etal-2023-instructscore}. However, these methods can suffer from biases, hallucinations, and prompt sensitivity, leading to inconsistent results~\cite{panickssery2024llm,chiang-lee-2023-lm-evaluation,shen-etal-2023-human-level}.

\noindent\textbf{Lecture-based Academic Datasets:} Effective analysis of educational lectures requires datasets integrating visual aids, spoken language, and textual content. Existing lecture datasets such as \textbf{LectureBank}~\cite{lecturebank}, \textbf{LectureVideoDB}~\cite{lecturevideodb}, and \textbf{GoogleI/O}~\cite{googleio} often lack comprehensive multimodal alignment, focus narrowly on specific tasks (e.g., concept mapping, summarization, engagement analysis, text recognition). While datasets like \textbf{LPM}~\cite{Lee2023} have advanced cross-modal integration of slide visuals and speech, they lackmetrics or annotations specifically for assessing educational quality.\looseness=-1


\section{\rewardmodel}

\textbf{Problem Formulation.} Given a slide $S$, we generate explanatory text $T_{S} = \xi(S,\mathcal{K})=\xi(S,\Phi(S))$, where $\mathcal{K} = \Phi(S)$ represents external knowledge extracted by $\Phi$. We evaluate $T_S$ using a metric $\mathcal{A}(S,T_S)=\mathcal{A}(S,\xi(S,\mathcal{K}))$ that assesses knowledge acquisition based on content alignment and enhancement.

\subsection{Data Construction}

Grounded in Mayer's principles of multimedia learning~\cite{multimedia-mayer}, our metric evaluates slide presentations on four key dimensions: \emph{Content Relevance (CR)}, \emph{Expressive Clarity (EC)}, \emph{Logical Structure (LS)}, and \emph{Audience Engagement (AE)}. Table~\ref{tab:criteria_principles} details these rubrics and their theoretical basis, ensuring a pedagogically meaningful assessment focused on optimizing comprehension and retention through effective integration of verbal and visual information.

\begin{table}[ht]
  \small
  \centering
    \caption{Rubrics design and supporting theoretical principles.}
    \begin{tabular}{lll}
        \toprule
        \textbf{Rub.} & \textbf{Principle} & \textbf{Description} \\
        \midrule
        CR & Modality Prin. & Verbal-visual content alignment. \\
        EC & Coherence Prin. & Explanation clarity and conciseness. \\
        LS & Segmenting Prin. & Logical organization of content. \\
        AE & Personalization Prin. & Audience connection and motivation. \\
        \bottomrule
    \end{tabular}
    \label{tab:criteria_principles}
\end{table}

\subsubsection{Heterogeneous Data Integration}

We utilize the \textbf{Lecture Presentations Multimodal (LPM)} dataset~\cite{Lee2023}, a diverse repository of academic lecture slides paired with corresponding transcripts. Leveraging its time-stamped video links and speech-transcript time-range annotations, we extract slide images and transcribed the accompanying audio. All transcriptions are manually reviewed to ensure accuracy and coherence.


\subsubsection{Multimodal Alignment and Refinement}

We manually align transcriptions with corresponding slides to ensure verbal explanations complement visual content. While raw transcriptions often suffer from speech disfluencies (e.g., fillers, errors, missing punctuation), hindering readability, we refine these transcriptions—removing disfluencies, correcting punctuation, and improving structure—while preserving the original speaker's intent. This process enhances clarity for subsequent annotation. All refinements undergo manual review to ensure fidelity.

\begin{table}[h]
  \centering
  \small
  \caption{Annotation Statistics: Inter-Annotator Agreement (IAA, Krippendorff's $\alpha$) and Exact Agreement Rate ($\phi$).}
  \begin{tabular}{lcc|cc}
        \toprule
         \textbf{Rubrics} & \textbf{Mean} & \textbf{Std} & $\alpha$ & $\phi$  \\
        \midrule
         Content Revelance (CR) & $3.88$ & $0.98$ & $0.84$ & $0.78$ \\
         Expressive Clarity (EC) & $4.14$ & $0.72$ & $0.75$ & $0.77$ \\
         Logical Structure (LS) & $4.34$ & $0.73$ & $0.68$ & $0.67$ \\
         Audience Engagement (AE) & $2.86$ & $0.73$ & $0.57$ & $0.64$ \\
        \bottomrule
  \end{tabular}
  \label{tab:annotation}
\end{table}


\subsubsection{Fine-grained Slide Presentation Assessment}

We employ three expert annotators, who are proficient in English, multimedia learning, and relevant subjects, to evaluate the slide presentations. To ensure consistency, we train the annotators on 5\% of the dataset. Inter-Annotator Agreement (IAA), measured via Krippendorff’s alpha~\cite{krippendorff2011computing}, is iteratively calibrated until a consensus is met. Each sample is independently rated 1-5 by the three annotators after reviewing the slide and text. We retain only samples meeting a high IAA threshold to ensure reliable ratings, as shown in Table~\ref{tab:annotation}.


\subsection{Data Analysis}

\subsubsection{Dataset Statistics}

Our curated dataset includes a collection of 56 lectures, encompassing a total of 2,097 samples. Each sample includes a slide image paired with a refined caption and ratings assigned by annotators based on our established evaluation rubrics. We offer this dataset as a valuable and robust resource for future research and applications in the field of image captioning, especially within the domain of educational presentation.


\subsubsection{Comparison with other datasets}

Table~\ref{tab:comparison} presents a comparative analysis of our dataset alongside other academic lecture-based datasets. While existing datasets primarily focus on multimodal recognition and content understanding, they often lack a structured framework for evaluating the effectiveness of slide-based educational materials. In contrast, our dataset is the first to explicitly support the assessment of slide presentations by integrating both manually processed slide segments and speech transcriptions. Furthermore, it features fine-grained human annotations aligned with our predefined scoring rubrics, enabling a more comprehensive and reliable evaluation of multimodal learning quality.

\begin{table}[ht]
  \centering
  \caption{Comparison with existing lecture-based datasets. \rewardmodel is the first to use manual slide \textit{Seg}mentation and \textit{Tran}scription supporting both \textit{Gen}eration and \textit{Eval}uation tasks. (A=Automated; M=Manual/partially manual)}
  \resizebox{0.95\linewidth}{!}{
    \begin{tabular}{l|ccc|cc|cc|c}
    \toprule
    \multirow{2}{*}{\textbf{Dataset}} & \multicolumn{3}{c|}{\textbf{Statistics}} & \multicolumn{2}{c|}{\textbf{Processing}} & \multicolumn{2}{c|}{\textbf{Tasks}} & \multirow{2}{*}{\textbf{Avai.}} \\
    \cmidrule(lr){2-8}
      & \multicolumn{1}{c}{\# Videos} & \multicolumn{1}{c}{\# Hours} & \multicolumn{1}{c|}{\# Slides} & \multicolumn{1}{c}{Seg.} & \multicolumn{1}{c|}{Tran.} & \multicolumn{1}{c}{Gen.} & \multicolumn{1}{c|}{Eval.} & \\
    \midrule
    LaRo.~\cite{larochelle} & $47$ & $65$ & $3250$ & A & A & \ding{51} & \ding{55} & \ding{55} \\
    LVDB~\cite{lecturevideodb} & $24$ & $-$ & $5474$ & M & $-$ & \ding{51} & \ding{55} & \ding{51} \\
    GoogleIO~\cite{googleio} & $209$ & $-$ & $-$ & A & A & \ding{51} & \ding{55} & \ding{51} \\
    LecB.~\cite{lecturebank} & $1352$ & $-$ & $51939$ & M & $-$ & \ding{51} & \ding{55} & \ding{51} \\
    ALV~\cite{alv} & $-$ & $-$ & $1498$ & A & A & \ding{51} & \ding{55} & \ding{51} \\
    VLEng.~\cite{vlengagement} & $11568$ & $-$ & $-$ & $-$ & $-$ & \ding{51} & \ding{55} & \ding{51} \\
    AVIATE~\cite{aviate} & $8201$ & $2300$ & $-$ & $-$ & A & \ding{51} & \ding{55} & \ding{51} \\
    LPM~\cite{Lee2023} & $334$ & $187$ & $9031$ & M & A & \ding{51} & \ding{55} & \ding{51} \\
    \midrule
    \rewardmodel (Ours) & $56$ & $65$ & $2097$ & M & M & \ding{51} & \ding{51} & \ding{51} \\
    \bottomrule
  \end{tabular}
  }
  \label{tab:comparison}
\end{table}

\subsection{Availability}

To support research and broader use, our dataset and associated resources are publicly available, including: (1) \textbf{Dataset}. A collection of $50+$ lectures, encompassing a total of $2000+$ samples. (2) \textbf{Data Processing Toolkit}. Querying functions and processing scripts for data retrieval, flexible structuring, and extending the methodology to other domains. (3) \textbf{Trained Reward Model}. A model trained on our dataset, demonstrating superior performance over baselines and strong alignment with human judgment (details in Section 4.3).

\begin{table*}[ht]
  \centering
  \small
  \setlength{\tabcolsep}{4pt}
  \caption{Pearson ($\sigma$), Spearman ($\rho$), Kendall-Tau ($\tau$) correlations between automated metrics and human evaluations. Metrics grouped into: (1) Reference-based (comparing generated vs. reference text), (2) Prompt-based LLM Evaluators (using prompted LLMs), and (3) \rewardmodel (our trained model, achieving the highest correlation).}
  \begin{tabular}{l|ccc|ccc|ccc|ccc|ccc}
    \toprule
    \multirow{2}{*}{\textbf{Metrics}} & \multicolumn{3}{c|}{\textbf{Content Relevance}} & \multicolumn{3}{c|}{\textbf{Expressive Clarity}} & \multicolumn{3}{c|}{\textbf{Logical Structure}} & \multicolumn{3}{c|}{\textbf{Audience Engagement}} & \multicolumn{3}{c}{\textbf{Overall}} \\
      & \multicolumn{1}{c}{$\sigma$} & \multicolumn{1}{c}{$\rho$} & \multicolumn{1}{c|}{$\tau$} & \multicolumn{1}{c}{$\sigma$} & \multicolumn{1}{c}{$\rho$} & \multicolumn{1}{c|}{$\tau$} & \multicolumn{1}{c}{$\sigma$} & \multicolumn{1}{c}{$\rho$} & \multicolumn{1}{c|}{$\tau$} & \multicolumn{1}{c}{$\sigma$} & \multicolumn{1}{c}{$\rho$} & \multicolumn{1}{c|}{$\tau$} & \multicolumn{1}{c}{$\sigma$} & \multicolumn{1}{c}{$\rho$} & \multicolumn{1}{c}{$\tau$} \\
    \midrule
    BLEU-4~\cite{bleu2002} & $0.11$ & $0.12$ & $0.12$ & $0.09$ & $0.10$ & $0.10$ & $0.10$ & $0.11$ & $0.10$ & $0.12$ & $0.18$ & $0.17$ & $0.11$ & $0.13$ & $0.12$ \\
    ROUGE-L~\cite{lin-2004-rouge} & $0.30$ & $0.24$ & $0.22$ & $0.30$ & $0.29$ & $0.25$ & $0.15$ & $0.15$ & $0.13$ & $0.06$ & $0.07$ & $0.06$ & $0.20$ & $0.19$ & $0.17$ \\
    METEOR~\cite{banerjee2005meteor} & $0.18$ & $0.17$ & $0.16$ & $0.07$ & $0.07$ & $0.07$ & $0.14$ & $0.11$ & $0.10$ & $-0.16$ & $-0.17$ & $-0.16$ & $0.06$ & $0.04$ & $0.04$ \\
    SPICE~\cite{anderson2016spice} & $0.03$ & $0.01$ & $0.01$ & $-0.07$ & $-0.2$ & $-0.18$ & $0.13$ & $0.12$ & $0.11$ & $0.14$ & $0.15$ & $0.14$ & $0.06$ & $0.02$ & $0.02$ \\
    CIDEr~\cite{vedantam2015cider} & $0.08$ & $0.10$ & $0.10$ & $0.10$ & $0.04$ & $0.03$ & $0.12$ & $0.15$ & $0.14$ & $0.12$ & $0.06$ & $0.06$ & $0.10$ & $0.09$ & $0.08$ \\
    \midrule
    GPT-4V~\cite{openai2023gpt} & $0.38$ & $0.29$ & $0.26$ & $0.26$ & $0.20$ & $0.18$ & $0.30$ & $0.27$ & $0.25$ & $0.34$ & $0.32$ & $0.29$ & $0.32$ & $0.27$ & $0.24$ \\
    GLM-4V~\cite{glm2024chatglm} & $0.01$ & $0.02$ & $0.01$ & $0.02$ & $0.02$ & $0.02$ & -$0.05$ & -$0.05$ & -$0.05$ & -$0.01$ & -$0.01$ & -$0.01$ & -$0.01$ & -$0.01$ & -$0.01$ \\
    MiniCPM-V2.5~\cite{yao2024minicpm} & $0.1$ & $0.06$ & $0.04$ & $0.12$ & $0.14$ & $0.12$ & $0.16$ & $0.17$ & $0.15$ & $0.07$ & $0.10$ & $0.08$ & $0.12$ & $0.12$ & $0.10$  \\
    G-Eval~\cite{liu2023geval} & $0.22$ & $0.26$ & $0.24$ & -$0.10$ & -$0.05$ & -$0.05$ & $0.12$ & $0.13$ & $0.11$ & $0.03$ & $0.04$ & $0.04$ & $0.07$ & $0.09$ & $0.08$ \\
    \midrule
    \rewardmodel (Ours) & $\textbf{0.74}$ & $\textbf{0.65}$ & $\textbf{0.58}$ & $\textbf{0.86}$ & $\textbf{0.84}$ & $\textbf{0.78}$ & $\textbf{0.78}$ & $\textbf{0.80}$ & $\textbf{0.72}$ & $\textbf{0.79}$ & $\textbf{0.79}$ & $\textbf{0.71}$ & $\textbf{0.79}$ & $\textbf{0.77}$ & $\textbf{0.70}$ \\
    \bottomrule
  \end{tabular}
  \label{tab:spearman_correlations}
\end{table*}

\section{EXPERIMENT}

\subsection{Main Experiment}
We evaluate our metric's effectiveness using a dataset of $420$ slide-text pairs. We measure \textbf{Pearson}, \textbf{Spearman}, and \textbf{Kendall-Tau} correlations between automated metrics and average human evaluations, which serve as our gold standard.

\subsubsection{Setup}

(1) \textbf{Reference-based Metrics:} We select five classical rule-based metrics as baselines: \emph{BLEU-4}~\cite{bleu2002}, \emph{ROUGE-L}~\cite{lin-2004-rouge}, \emph{METEOR}~\cite{banerjee2005meteor}, \emph{CIDEr}~\cite{vedantam2015cider}, and \emph{SPICE}~\cite{anderson2016spice}. 
We use instructor-prepared lecture materials (slides and presentations) as gold-standard references and generate explanatory text using a predefined prompt. We then apply these reference-based metrics to compare the generated text against the human reference.
(2) \textbf{Prompt-based LLM Evaluators:} Leveraging LLMs' ability to process multimodal information and evaluate content, we use prominent models as reference-free LLM-based evaluation baselines: \emph{GPT-4V}~\cite{openai2023gpt}, \emph{GLM-4V}~\cite{glm2024chatglm}, and \emph{MiniCPM-V2.5}~\cite{yao2024minicpm}. 
We also include \emph{G-Eval}~\cite{liu2023geval} as a baseline, utilizing its automated CoT prompts and structured framework for consistent evaluation.
(3) \textbf{\rewardmodel:} We employ the $8$B parameter \emph{MiniCPM-V2.5}~\cite{yao2024minicpm} as our backbone model, chosen for its strong multimodal capabilities and efficiency. This model is fine-tuned on our curated dataset to serve as our proposed \rewardmodel.

\subsubsection{Results}

Table~\ref{tab:spearman_correlations} presents the correlation results between the automated metrics and human evaluations.
(1) \textbf{Reference-based Metrics:} The results indicate that classical reference-based metrics exhibit notable biases compared to human ratings. While these metrics effectively measure content similarity to reference texts, they fail to account for qualitative aspects such as critical insights or novel ideas that diverge from the reference material. This limitation underscores the broader challenge of reference-based metrics in contexts requiring more sophisticated evaluations.
(2) \textbf{Prompt-based LLM Evaluators:} While capable, these LLMs show limitations compared to human assessment. Their correlations with human judgments are generally low, indicating issues such as evaluation gaps, poor score differentiation, and biases towards verbosity and self-generated content, which undermine their reliability and fairness. This highlights the need for refinement beyond simple prompting.
(3) \textbf{\rewardmodel:} Our fine-tuned model, \rewardmodel, significantly outperforms both reference-based and prompt-based baselines across all criteria and overall. Its correlations approach human inter-annotator agreement levels. Specifically, the substantial performance gain over its vanilla version (MiniCPM-V2.5, with correlations around $0.1$) further validates the effectiveness of our fine-tuning approach and the quality of our curated dataset.

\subsection{Ablation Study}

To gain deeper insights into the factors that influence \rewardmodel's performance, we conduct a series of ablation studies to examine key aspects of the model, including (1) \textit{data scalability}, (2) \textit{scoring functions}, and (3) \textit{evaluation strategies}. 

\subsubsection{Data Scalability}

\begin{figure}[ht]
  \small
  \centering
    \includegraphics[width=0.8\linewidth]{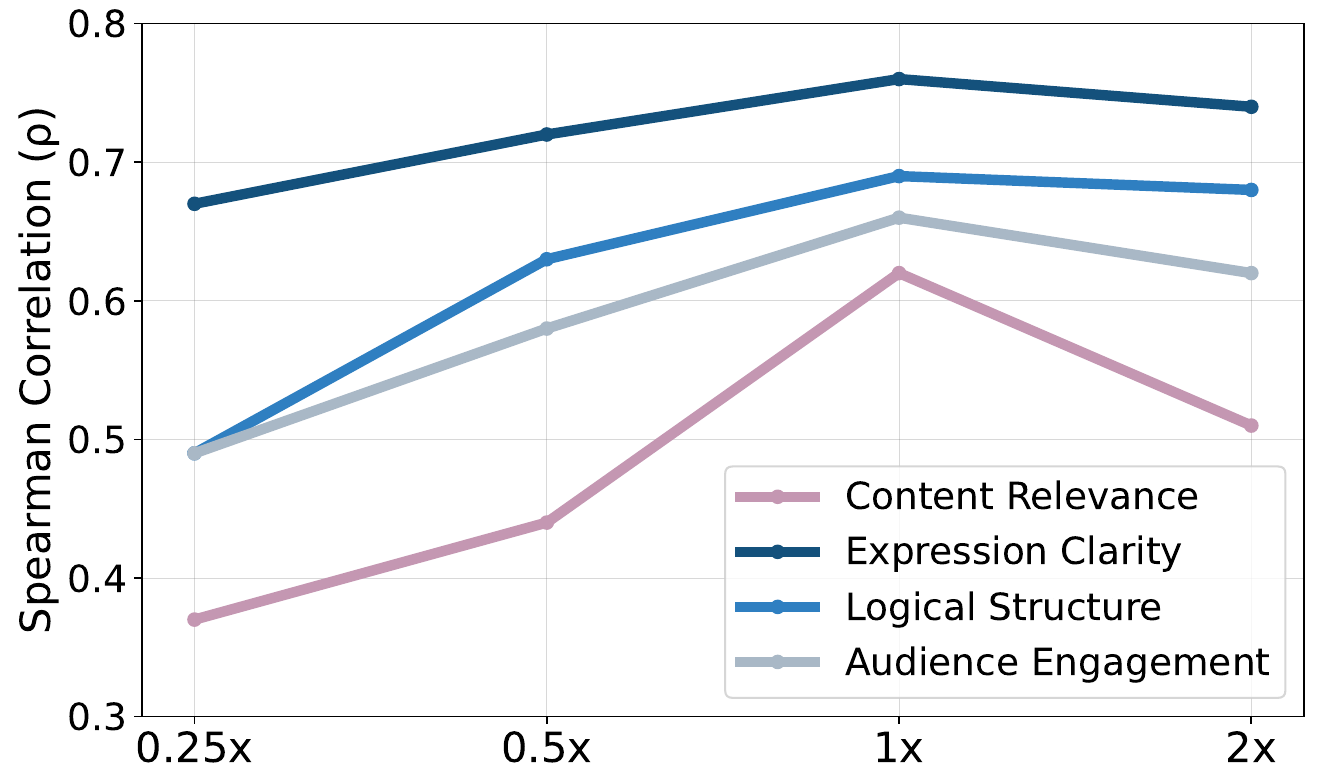}
    \caption{Spearman correlations ($\rho$) between training data scalability and model's performance.}
    \label{fig:ablation-data-size}
  \end{figure}

We examine data scalability by adjusting training data size. As shown in Figure~\ref{fig:ablation-data-size}, performance initially improves with more data, underscoring the need for sufficient training samples. However, doubling the data led to overfitting noise and reduced generalization. Optimizing data volume is thus essential to balance model generalization and performance, preventing overfitting while maintaining robust evaluations.


\subsubsection{Scoring Function}

We investigate if a continuous scoring function could offer finer granularity than discrete ratings (1-5). This function computes the expected score based on the model's output probabilities for each discrete score: $\mathcal{A}_{norm}(S, C_S) = \sum_{i=1}^{5} p(s_i) \times s_i$, where $p(s_i)$ is the probability of discrete score $s_i \in \{1,2,3,4,5\}$. As shown in Figure~\ref{fig:ablation-continual-distinct}, the continuous scores closely align with the discrete ones. This occurs because the model typically assigns a probability near 1 to a single discrete score, indicating high confidence. We conclude that the discrete 1-5 scale provides sufficient granularity for evaluating slide presentations on specific rubrics.

\subsubsection{Criterion-Specific Modeling}

\begin{figure}[ht]
  \centering
    \includegraphics[width=\columnwidth]{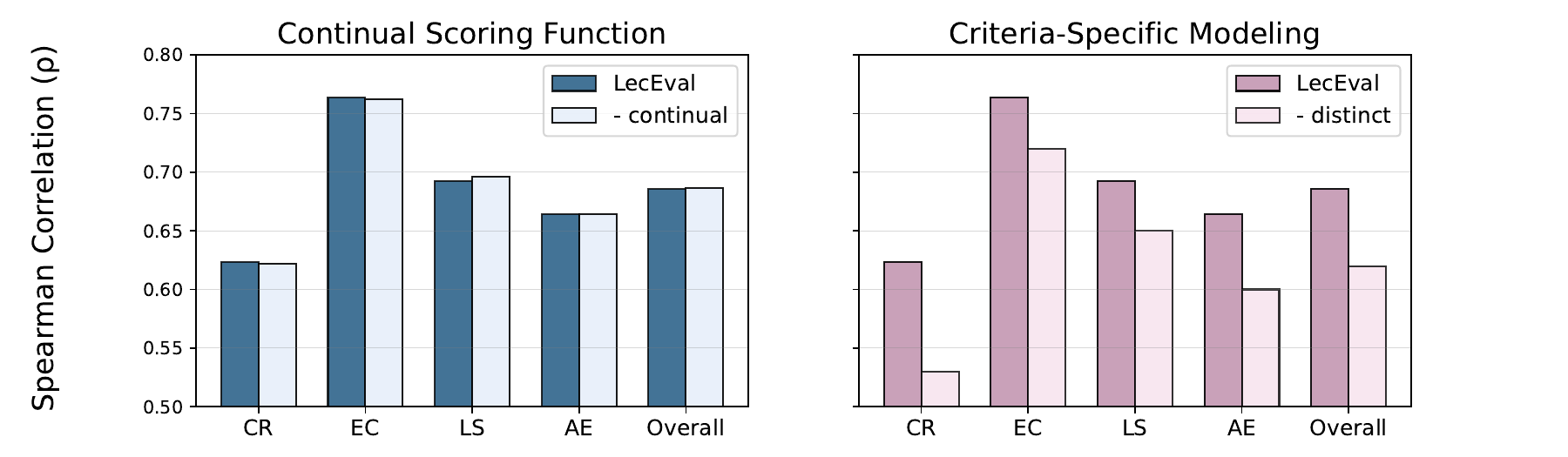}
    \caption{Correlations between (1) scoring function, and (2) criterion-specific modeling with model's performance.}
    \label{fig:ablation-continual-distinct}
  \end{figure}

We investigate whether training separate models for each criterion improves performance over a single integrated model by enabling targeted assessments. Contrary to expectations, Figure~\ref{fig:ablation-continual-distinct} shows that these criterion-specific models performed worse across all criteria compared to the integrated model. This suggests the integrated model benefits from leveraging interactions between criteria for a more holistic assessment, highlighting the advantage of a unified evaluation approach over isolated, criterion-specific training.

\section{CONCLUSION}

In this paper, we address the challenge of evaluating slide-based learning effectiveness by introducing \rewardmodel, a novel metric for assessing multimodal knowledge acquisition. Grounded in Mayer’s Cognitive Theory of Multimedia Learning, \rewardmodel employs four key rubrics: \emph{Content Relevance}, \emph{Expressive Clarity}, \emph{Logical Structure}, and \emph{Audience Engagement}, designed to reflect core multimedia learning principles integrating verbal and visual information.
To develop and validate our proposed method, we curate a dataset of over $2,000$ slide-text pairs from $50+$ real online lectures, featuring fine-grained human annotations. Our experiments demonstrate that \rewardmodel accurately captures human evaluative nuances and significantly outperforms both reference-based and reference-free baselines in accuracy and reliability. 
By translating foundational multimedia learning principles into a scalable, automated assessment tool, \rewardmodel represents a significant advancement in the data-driven evaluation of multimodal educational content.

\clearpage
\bibliographystyle{ACM-Reference-Format}
\bibliography{sample-base}


\begin{thebibliography}{68}


\ifx \showCODEN    \undefined \def \showCODEN     #1{\unskip}     \fi
\ifx \showISBNx    \undefined \def \showISBNx     #1{\unskip}     \fi
\ifx \showISBNxiii \undefined \def \showISBNxiii  #1{\unskip}     \fi
\ifx \showISSN     \undefined \def \showISSN      #1{\unskip}     \fi
\ifx \showLCCN     \undefined \def \showLCCN      #1{\unskip}     \fi
\ifx \shownote     \undefined \def \shownote      #1{#1}          \fi
\ifx \showarticletitle \undefined \def \showarticletitle #1{#1}   \fi
\ifx \showURL      \undefined \def \showURL       {\relax}        \fi
\providecommand\bibfield[2]{#2}
\providecommand\bibinfo[2]{#2}
\providecommand\natexlab[1]{#1}
\providecommand\showeprint[2][]{arXiv:#2}

\bibitem[Anderson et~al\mbox{.}(2016)]%
        {anderson2016spice}
\bibfield{author}{\bibinfo{person}{Peter Anderson}, \bibinfo{person}{Basura
  Fernando}, \bibinfo{person}{Mark Johnson}, {and} \bibinfo{person}{Stephen
  Gould}.} \bibinfo{year}{2016}\natexlab{}.
\newblock \showarticletitle{Spice: Semantic propositional image caption
  evaluation}. In \bibinfo{booktitle}{\emph{Computer Vision--ECCV 2016: 14th
  European Conference, Amsterdam, The Netherlands, October 11-14, 2016,
  Proceedings, Part V 14}}. Springer, \bibinfo{pages}{382--398}.
\newblock


\bibitem[Angrist et~al\mbox{.}(2024)]%
        {angrist2024improve}
\bibfield{author}{\bibinfo{person}{Noam Angrist}, \bibinfo{person}{David~K
  Evans}, \bibinfo{person}{Deon Filmer}, \bibinfo{person}{Rachel Glennerster},
  \bibinfo{person}{Halsey Rogers}, {and} \bibinfo{person}{Shwetlena Sabarwal}.}
  \bibinfo{year}{2024}\natexlab{}.
\newblock \showarticletitle{How to improve education outcomes most efficiently?
  A review of the evidence using a unified metric}.
\newblock \bibinfo{journal}{\emph{Journal of Development Economics}}
  (\bibinfo{year}{2024}), \bibinfo{pages}{103382}.
\newblock


\bibitem[Atri et~al\mbox{.}(2021)]%
        {aviate}
\bibfield{author}{\bibinfo{person}{Yash~Kumar Atri}, \bibinfo{person}{Shraman
  Pramanick}, \bibinfo{person}{Vikram Goyal}, {and} \bibinfo{person}{Tanmoy
  Chakraborty}.} \bibinfo{year}{2021}\natexlab{}.
\newblock \bibinfo{title}{See, Hear, Read: Leveraging Multimodality with Guided
  Attention for Abstractive Text Summarization}.
\newblock
\showeprint[arxiv]{2105.09601}~[cs.LG]


\bibitem[Bai et~al\mbox{.}(2024)]%
        {bai2024benchmarking}
\bibfield{author}{\bibinfo{person}{Yushi Bai}, \bibinfo{person}{Jiahao Ying},
  \bibinfo{person}{Yixin Cao}, \bibinfo{person}{Xin Lv}, \bibinfo{person}{Yuze
  He}, \bibinfo{person}{Xiaozhi Wang}, \bibinfo{person}{Jifan Yu},
  \bibinfo{person}{Kaisheng Zeng}, \bibinfo{person}{Yijia Xiao},
  \bibinfo{person}{Haozhe Lyu}, {et~al\mbox{.}}}
  \bibinfo{year}{2024}\natexlab{}.
\newblock \showarticletitle{Benchmarking foundation models with
  language-model-as-an-examiner}.
\newblock \bibinfo{journal}{\emph{Advances in Neural Information Processing
  Systems}}  \bibinfo{volume}{36} (\bibinfo{year}{2024}).
\newblock


\bibitem[Banerjee and Lavie(2005)]%
        {banerjee2005meteor}
\bibfield{author}{\bibinfo{person}{Satanjeev Banerjee} {and}
  \bibinfo{person}{Alon Lavie}.} \bibinfo{year}{2005}\natexlab{}.
\newblock \showarticletitle{METEOR: An automatic metric for MT evaluation with
  improved correlation with human judgments}. In
  \bibinfo{booktitle}{\emph{Proceedings of the acl workshop on intrinsic and
  extrinsic evaluation measures for machine translation and/or summarization}}.
  \bibinfo{pages}{65--72}.
\newblock


\bibitem[Bernardi et~al\mbox{.}(2016)]%
        {bernardi2016automatic}
\bibfield{author}{\bibinfo{person}{Raffaella Bernardi}, \bibinfo{person}{Ruket
  Cakici}, \bibinfo{person}{Desmond Elliott}, \bibinfo{person}{Aykut Erdem},
  \bibinfo{person}{Erkut Erdem}, \bibinfo{person}{Nazli Ikizler-Cinbis},
  \bibinfo{person}{Frank Keller}, \bibinfo{person}{Adrian Muscat}, {and}
  \bibinfo{person}{Barbara Plank}.} \bibinfo{year}{2016}\natexlab{}.
\newblock \showarticletitle{Automatic description generation from images: A
  survey of models, datasets, and evaluation measures}.
\newblock \bibinfo{journal}{\emph{Journal of Artificial Intelligence Research}}
   \bibinfo{volume}{55} (\bibinfo{year}{2016}), \bibinfo{pages}{409--442}.
\newblock


\bibitem[Brock and Joglekar(2011)]%
        {brock2011empowering}
\bibfield{author}{\bibinfo{person}{Sabra~E Brock} {and} \bibinfo{person}{Yogini
  Joglekar}.} \bibinfo{year}{2011}\natexlab{}.
\newblock \showarticletitle{Empowering PowerPoint: Slides and teaching
  effectiveness}.
\newblock \bibinfo{journal}{\emph{Interdisciplinary Journal of Information,
  Knowledge, and Management}}  \bibinfo{volume}{6} (\bibinfo{year}{2011}),
  \bibinfo{pages}{85}.
\newblock


\bibitem[Buckingham(2007)]%
        {buckingham2007digital}
\bibfield{author}{\bibinfo{person}{David Buckingham}.}
  \bibinfo{year}{2007}\natexlab{}.
\newblock \showarticletitle{Digital Media Literacies: rethinking media
  education in the age of the Internet}.
\newblock \bibinfo{journal}{\emph{Research in comparative and international
  education}} \bibinfo{volume}{2}, \bibinfo{number}{1} (\bibinfo{year}{2007}),
  \bibinfo{pages}{43--55}.
\newblock


\bibitem[Buckingham(2013)]%
        {buckingham2013media}
\bibfield{author}{\bibinfo{person}{David Buckingham}.}
  \bibinfo{year}{2013}\natexlab{}.
\newblock \bibinfo{booktitle}{\emph{Media education: Literacy, learning and
  contemporary culture}}.
\newblock \bibinfo{publisher}{John Wiley \& Sons}.
\newblock


\bibitem[Bulathwela et~al\mbox{.}(2020)]%
        {vlengagement}
\bibfield{author}{\bibinfo{person}{Sahan Bulathwela}, \bibinfo{person}{Maria
  Perez-Ortiz}, \bibinfo{person}{Emine Yilmaz}, {and} \bibinfo{person}{John
  Shawe-Taylor}.} \bibinfo{year}{2020}\natexlab{}.
\newblock \bibinfo{title}{VLEngagement: A Dataset of Scientific Video Lectures
  for Evaluating Population-based Engagement}.
\newblock
\showeprint[arxiv]{2011.02273}~[cs.CY]
\urldef\tempurl%
\url{https://arxiv.org/abs/2011.02273}
\showURL{%
\tempurl}


\bibitem[Chen et~al\mbox{.}(2014)]%
        {googleio}
\bibfield{author}{\bibinfo{person}{Huizhong Chen}, \bibinfo{person}{Matthew
  Cooper}, \bibinfo{person}{Dhiraj Joshi}, {and} \bibinfo{person}{Bernd
  Girod}.} \bibinfo{year}{2014}\natexlab{}.
\newblock \showarticletitle{Multi-modal Language Models for Lecture Video
  Retrieval}. In \bibinfo{booktitle}{\emph{Proceedings of the 22nd ACM
  International Conference on Multimedia}} (Orlando, Florida, USA)
  \emph{(\bibinfo{series}{MM '14})}. \bibinfo{publisher}{Association for
  Computing Machinery}, \bibinfo{address}{New York, NY, USA},
  \bibinfo{pages}{1081–1084}.
\newblock
\showISBNx{9781450330633}
\href{https://doi.org/10.1145/2647868.2654964}{doi:\nolinkurl{10.1145/2647868.2654964}}


\bibitem[Chiang and Lee(2023a)]%
        {chiang-lee-2023-large}
\bibfield{author}{\bibinfo{person}{Cheng-Han Chiang} {and}
  \bibinfo{person}{Hung-yi Lee}.} \bibinfo{year}{2023}\natexlab{a}.
\newblock \showarticletitle{Can Large Language Models Be an Alternative to
  Human Evaluations?}. In \bibinfo{booktitle}{\emph{Proceedings of the 61st
  Annual Meeting of the Association for Computational Linguistics (Volume 1:
  Long Papers)}}, \bibfield{editor}{\bibinfo{person}{Anna Rogers},
  \bibinfo{person}{Jordan Boyd-Graber}, {and} \bibinfo{person}{Naoaki Okazaki}}
  (Eds.). \bibinfo{publisher}{Association for Computational Linguistics},
  \bibinfo{address}{Toronto, Canada}, \bibinfo{pages}{15607--15631}.
\newblock
\href{https://doi.org/10.18653/v1/2023.acl-long.870}{doi:\nolinkurl{10.18653/v1/2023.acl-long.870}}


\bibitem[Chiang and Lee(2023b)]%
        {chiang-lee-2023-lm-evaluation}
\bibfield{author}{\bibinfo{person}{Cheng-Han Chiang} {and}
  \bibinfo{person}{Hung-yi Lee}.} \bibinfo{year}{2023}\natexlab{b}.
\newblock \showarticletitle{Can Large Language Models Be an Alternative to
  Human Evaluations?}. In \bibinfo{booktitle}{\emph{Proceedings of the 61st
  Annual Meeting of the Association for Computational Linguistics (Volume 1:
  Long Papers)}}, \bibfield{editor}{\bibinfo{person}{Anna Rogers},
  \bibinfo{person}{Jordan Boyd-Graber}, {and} \bibinfo{person}{Naoaki Okazaki}}
  (Eds.). \bibinfo{publisher}{Association for Computational Linguistics},
  \bibinfo{address}{Toronto, Canada}, \bibinfo{pages}{15607--15631}.
\newblock
\href{https://doi.org/10.18653/v1/2023.acl-long.870}{doi:\nolinkurl{10.18653/v1/2023.acl-long.870}}


\bibitem[Dutta et~al\mbox{.}(2018)]%
        {lecturevideodb}
\bibfield{author}{\bibinfo{person}{Kartik Dutta}, \bibinfo{person}{Minesh
  Mathew}, \bibinfo{person}{Praveen Krishnan}, {and} \bibinfo{person}{C.V
  Jawahar}.} \bibinfo{year}{2018}\natexlab{}.
\newblock \showarticletitle{Localizing and Recognizing Text in Lecture Videos}.
  In \bibinfo{booktitle}{\emph{2018 16th International Conference on Frontiers
  in Handwriting Recognition (ICFHR)}}. \bibinfo{pages}{235--240}.
\newblock
\href{https://doi.org/10.1109/ICFHR-2018.2018.00049}{doi:\nolinkurl{10.1109/ICFHR-2018.2018.00049}}


\bibitem[Fu et~al\mbox{.}(2023)]%
        {fu2023gptscore}
\bibfield{author}{\bibinfo{person}{Jinlan Fu}, \bibinfo{person}{See-Kiong Ng},
  \bibinfo{person}{Zhengbao Jiang}, {and} \bibinfo{person}{Pengfei Liu}.}
  \bibinfo{year}{2023}\natexlab{}.
\newblock \bibinfo{title}{GPTScore: Evaluate as You Desire}.
\newblock
\showeprint[arxiv]{2302.04166}~[cs.CL]


\bibitem[Galanopoulos and Mezaris(2019)]%
        {alv}
\bibfield{author}{\bibinfo{person}{Damianos Galanopoulos} {and}
  \bibinfo{person}{Vasileios Mezaris}.} \bibinfo{year}{2019}\natexlab{}.
\newblock \showarticletitle{Temporal Lecture Video Fragmentation Using Word
  Embeddings}. In \bibinfo{booktitle}{\emph{MultiMedia Modeling}},
  \bibfield{editor}{\bibinfo{person}{Ioannis Kompatsiaris},
  \bibinfo{person}{Benoit Huet}, \bibinfo{person}{Vasileios Mezaris},
  \bibinfo{person}{Cathal Gurrin}, \bibinfo{person}{Wen-Huang Cheng}, {and}
  \bibinfo{person}{Stefanos Vrochidis}} (Eds.). \bibinfo{publisher}{Springer
  International Publishing}, \bibinfo{address}{Cham},
  \bibinfo{pages}{254--265}.
\newblock
\showISBNx{978-3-030-05716-9}


\bibitem[Gao et~al\mbox{.}(2024)]%
        {gao2024llmbased}
\bibfield{author}{\bibinfo{person}{Mingqi Gao}, \bibinfo{person}{Xinyu Hu},
  \bibinfo{person}{Jie Ruan}, \bibinfo{person}{Xiao Pu}, {and}
  \bibinfo{person}{Xiaojun Wan}.} \bibinfo{year}{2024}\natexlab{}.
\newblock \bibinfo{title}{LLM-based NLG Evaluation: Current Status and
  Challenges}.
\newblock
\showeprint[arxiv]{2402.01383}~[cs.CL]


\bibitem[GLM(2024)]%
        {glm2024chatglm}
\bibfield{author}{\bibinfo{person}{Team GLM}.} \bibinfo{year}{2024}\natexlab{}.
\newblock \bibinfo{title}{ChatGLM: A Family of Large Language Models from
  GLM-130B to GLM-4 All Tools}.
\newblock
\showeprint[arxiv]{2406.12793}~[cs.CL]
\urldef\tempurl%
\url{https://arxiv.org/abs/2406.12793}
\showURL{%
\tempurl}


\bibitem[Graham(2015)]%
        {graham2015re}
\bibfield{author}{\bibinfo{person}{Yvette Graham}.}
  \bibinfo{year}{2015}\natexlab{}.
\newblock \showarticletitle{Re-evaluating automatic summarization with BLEU and
  192 shades of ROUGE}. In \bibinfo{booktitle}{\emph{Proceedings of the 2015
  conference on empirical methods in natural language processing}}.
  \bibinfo{pages}{128--137}.
\newblock


\bibitem[Hamdhana et~al\mbox{.}(2024)]%
        {hamdhana2024improved}
\bibfield{author}{\bibinfo{person}{Defry Hamdhana}, \bibinfo{person}{Haru
  Kaneko}, \bibinfo{person}{John~Noel Victorino}, {and} \bibinfo{person}{Sozo
  Inoue}.} \bibinfo{year}{2024}\natexlab{}.
\newblock \showarticletitle{Improved Evaluation Metrics for Sentence
  Suggestions in Nursing and Elderly Care Record Applications}. In
  \bibinfo{booktitle}{\emph{Healthcare}}, Vol.~\bibinfo{volume}{12}. MDPI,
  \bibinfo{pages}{367}.
\newblock


\bibitem[He et~al\mbox{.}(2008)]%
        {rouge-2}
\bibfield{author}{\bibinfo{person}{Tingting He}, \bibinfo{person}{Jinguang
  Chen}, \bibinfo{person}{Liang Ma}, \bibinfo{person}{Zhuoming Gui},
  \bibinfo{person}{Fang Li}, \bibinfo{person}{Wei Shao}, {and}
  \bibinfo{person}{Qian Wang}.} \bibinfo{year}{2008}\natexlab{}.
\newblock \showarticletitle{ROUGE-C: A fully automated evaluation method for
  multi-document summarization}. In \bibinfo{booktitle}{\emph{2008 IEEE
  International Conference on Granular Computing}}. \bibinfo{pages}{269--274}.
\newblock
\href{https://doi.org/10.1109/GRC.2008.4664680}{doi:\nolinkurl{10.1109/GRC.2008.4664680}}


\bibitem[Herdade et~al\mbox{.}(2019)]%
        {herdade2019image}
\bibfield{author}{\bibinfo{person}{Simao Herdade}, \bibinfo{person}{Armin
  Kappeler}, \bibinfo{person}{Kofi Boakye}, {and} \bibinfo{person}{Joao
  Soares}.} \bibinfo{year}{2019}\natexlab{}.
\newblock \showarticletitle{Image captioning: Transforming objects into words}.
\newblock \bibinfo{journal}{\emph{Advances in neural information processing
  systems}}  \bibinfo{volume}{32} (\bibinfo{year}{2019}).
\newblock


\bibitem[Hill and Grossman(2013)]%
        {hill2013learning}
\bibfield{author}{\bibinfo{person}{Heather Hill} {and} \bibinfo{person}{Pam
  Grossman}.} \bibinfo{year}{2013}\natexlab{}.
\newblock \showarticletitle{Learning from teacher observations: Challenges and
  opportunities posed by new teacher evaluation systems}.
\newblock \bibinfo{journal}{\emph{Harvard educational review}}
  \bibinfo{volume}{83}, \bibinfo{number}{2} (\bibinfo{year}{2013}),
  \bibinfo{pages}{371--384}.
\newblock


\bibitem[Jenkins(2009)]%
        {jenkins2009confronting}
\bibfield{author}{\bibinfo{person}{Henry Jenkins}.}
  \bibinfo{year}{2009}\natexlab{}.
\newblock \bibinfo{booktitle}{\emph{Confronting the challenges of participatory
  culture: Media education for the 21st century}}.
\newblock \bibinfo{publisher}{The MIT press}.
\newblock


\bibitem[Jia et~al\mbox{.}(2023)]%
        {jia-etal-2023-zero}
\bibfield{author}{\bibinfo{person}{Qi Jia}, \bibinfo{person}{Siyu Ren},
  \bibinfo{person}{Yizhu Liu}, {and} \bibinfo{person}{Kenny Zhu}.}
  \bibinfo{year}{2023}\natexlab{}.
\newblock \showarticletitle{Zero-shot Faithfulness Evaluation for Text
  Summarization with Foundation Language Model}. In
  \bibinfo{booktitle}{\emph{Proceedings of the 2023 Conference on Empirical
  Methods in Natural Language Processing}},
  \bibfield{editor}{\bibinfo{person}{Houda Bouamor}, \bibinfo{person}{Juan
  Pino}, {and} \bibinfo{person}{Kalika Bali}} (Eds.).
  \bibinfo{publisher}{Association for Computational Linguistics},
  \bibinfo{address}{Singapore}, \bibinfo{pages}{11017--11031}.
\newblock
\href{https://doi.org/10.18653/v1/2023.emnlp-main.679}{doi:\nolinkurl{10.18653/v1/2023.emnlp-main.679}}


\bibitem[Jiang et~al\mbox{.}(2024)]%
        {jiang2024tigerscore}
\bibfield{author}{\bibinfo{person}{Dongfu Jiang}, \bibinfo{person}{Yishan Li},
  \bibinfo{person}{Ge Zhang}, \bibinfo{person}{Wenhao Huang},
  \bibinfo{person}{Bill~Yuchen Lin}, {and} \bibinfo{person}{Wenhu Chen}.}
  \bibinfo{year}{2024}\natexlab{}.
\newblock \bibinfo{title}{TIGERScore: Towards Building Explainable Metric for
  All Text Generation Tasks}.
\newblock
\showeprint[arxiv]{2310.00752}~[cs.CL]
\urldef\tempurl%
\url{https://arxiv.org/abs/2310.00752}
\showURL{%
\tempurl}


\bibitem[Ke et~al\mbox{.}(2024)]%
        {ke-etal-2024-critiquellm}
\bibfield{author}{\bibinfo{person}{Pei Ke}, \bibinfo{person}{Bosi Wen},
  \bibinfo{person}{Andrew Feng}, \bibinfo{person}{Xiao Liu},
  \bibinfo{person}{Xuanyu Lei}, \bibinfo{person}{Jiale Cheng},
  \bibinfo{person}{Shengyuan Wang}, \bibinfo{person}{Aohan Zeng},
  \bibinfo{person}{Yuxiao Dong}, \bibinfo{person}{Hongning Wang},
  \bibinfo{person}{Jie Tang}, {and} \bibinfo{person}{Minlie Huang}.}
  \bibinfo{year}{2024}\natexlab{}.
\newblock \showarticletitle{{C}ritique{LLM}: Towards an Informative Critique
  Generation Model for Evaluation of Large Language Model Generation}. In
  \bibinfo{booktitle}{\emph{Proceedings of the 62nd Annual Meeting of the
  Association for Computational Linguistics (Volume 1: Long Papers)}},
  \bibfield{editor}{\bibinfo{person}{Lun-Wei Ku}, \bibinfo{person}{Andre
  Martins}, {and} \bibinfo{person}{Vivek Srikumar}} (Eds.).
  \bibinfo{publisher}{Association for Computational Linguistics},
  \bibinfo{address}{Bangkok, Thailand}, \bibinfo{pages}{13034--13054}.
\newblock
\urldef\tempurl%
\url{https://aclanthology.org/2024.acl-long.704}
\showURL{%
\tempurl}


\bibitem[Kilickaya et~al\mbox{.}(2016)]%
        {kilickaya2016}
\bibfield{author}{\bibinfo{person}{Mert Kilickaya}, \bibinfo{person}{Aykut
  Erdem}, \bibinfo{person}{Nazli Ikizler-Cinbis}, {and} \bibinfo{person}{Erkut
  Erdem}.} \bibinfo{year}{2016}\natexlab{}.
\newblock \bibinfo{title}{Re-evaluating Automatic Metrics for Image
  Captioning}.
\newblock
\showeprint[arxiv]{1612.07600}~[cs.CL]
\urldef\tempurl%
\url{https://arxiv.org/abs/1612.07600}
\showURL{%
\tempurl}


\bibitem[Kim et~al\mbox{.}(2024)]%
        {kim2024prometheus}
\bibfield{author}{\bibinfo{person}{Seungone Kim}, \bibinfo{person}{Jamin Shin},
  \bibinfo{person}{Yejin Cho}, \bibinfo{person}{Joel Jang},
  \bibinfo{person}{Shayne Longpre}, \bibinfo{person}{Hwaran Lee},
  \bibinfo{person}{Sangdoo Yun}, \bibinfo{person}{Seongjin Shin},
  \bibinfo{person}{Sungdong Kim}, \bibinfo{person}{James Thorne}, {and}
  \bibinfo{person}{Minjoon Seo}.} \bibinfo{year}{2024}\natexlab{}.
\newblock \bibinfo{title}{Prometheus: Inducing Fine-grained Evaluation
  Capability in Language Models}.
\newblock
\showeprint[arxiv]{2310.08491}~[cs.CL]


\bibitem[Krippendorff(2011)]%
        {krippendorff2011computing}
\bibfield{author}{\bibinfo{person}{Klaus Krippendorff}.}
  \bibinfo{year}{2011}\natexlab{}.
\newblock \bibinfo{title}{Computing Krippendorff’s alpha-reliability}.
\newblock


\bibitem[Lee et~al\mbox{.}(2023)]%
        {Lee2023}
\bibfield{author}{\bibinfo{person}{Dong~Won Lee}, \bibinfo{person}{Chaitanya
  Ahuja}, \bibinfo{person}{Paul~Pu Liang}, \bibinfo{person}{Sanika Natu}, {and}
  \bibinfo{person}{Louis-Philippe Morency}.} \bibinfo{year}{2023}\natexlab{}.
\newblock \showarticletitle{Lecture Presentations Multimodal Dataset: Towards
  Understanding Multimodality in Educational Videos}. In
  \bibinfo{booktitle}{\emph{2023 IEEE/CVF International Conference on Computer
  Vision (ICCV)}}. \bibinfo{pages}{20030--20041}.
\newblock
\href{https://doi.org/10.1109/ICCV51070.2023.01838}{doi:\nolinkurl{10.1109/ICCV51070.2023.01838}}


\bibitem[Le{\'o}n and Garc{\'\i}a-Mart{\'\i}nez(2021)]%
        {leon2021impact}
\bibfield{author}{\bibinfo{person}{Samuel~P Le{\'o}n} {and}
  \bibinfo{person}{Inmaculada Garc{\'\i}a-Mart{\'\i}nez}.}
  \bibinfo{year}{2021}\natexlab{}.
\newblock \showarticletitle{Impact of the provision of PowerPoint slides on
  learning}.
\newblock \bibinfo{journal}{\emph{Computers \& Education}}
  \bibinfo{volume}{173} (\bibinfo{year}{2021}), \bibinfo{pages}{104283}.
\newblock


\bibitem[Li et~al\mbox{.}(2019)]%
        {lecturebank}
\bibfield{author}{\bibinfo{person}{Irene Li}, \bibinfo{person}{Alexander~R.
  Fabbri}, \bibinfo{person}{Robert~R. Tung}, {and} \bibinfo{person}{Dragomir~R.
  Radev}.} \bibinfo{year}{2019}\natexlab{}.
\newblock \showarticletitle{What should I learn first: introducing LectureBank
  for NLP education and prerequisite chain learning}. In
  \bibinfo{booktitle}{\emph{Proceedings of the Thirty-Third AAAI Conference on
  Artificial Intelligence and Thirty-First Innovative Applications of
  Artificial Intelligence Conference and Ninth AAAI Symposium on Educational
  Advances in Artificial Intelligence}} (Honolulu, Hawaii, USA)
  \emph{(\bibinfo{series}{AAAI'19/IAAI'19/EAAI'19})}. \bibinfo{publisher}{AAAI
  Press}, Article \bibinfo{articleno}{819}, \bibinfo{numpages}{8}~pages.
\newblock
\showISBNx{978-1-57735-809-1}
\href{https://doi.org/10.1609/aaai.v33i01.33016674}{doi:\nolinkurl{10.1609/aaai.v33i01.33016674}}


\bibitem[Lin(2004)]%
        {lin-2004-rouge}
\bibfield{author}{\bibinfo{person}{Chin-Yew Lin}.}
  \bibinfo{year}{2004}\natexlab{}.
\newblock \showarticletitle{{ROUGE}: A Package for Automatic Evaluation of
  Summaries}. In \bibinfo{booktitle}{\emph{Text Summarization Branches Out}}.
  \bibinfo{publisher}{Association for Computational Linguistics},
  \bibinfo{address}{Barcelona, Spain}, \bibinfo{pages}{74--81}.
\newblock
\urldef\tempurl%
\url{https://aclanthology.org/W04-1013}
\showURL{%
\tempurl}


\bibitem[Liu and Liu(2010)]%
        {rouge-weakness-1}
\bibfield{author}{\bibinfo{person}{Feifan Liu} {and} \bibinfo{person}{Yang
  Liu}.} \bibinfo{year}{2010}\natexlab{}.
\newblock \showarticletitle{Exploring Correlation Between ROUGE and Human
  Evaluation on Meeting Summaries}.
\newblock \bibinfo{journal}{\emph{IEEE Transactions on Audio, Speech, and
  Language Processing}} \bibinfo{volume}{18}, \bibinfo{number}{1}
  (\bibinfo{year}{2010}), \bibinfo{pages}{187--196}.
\newblock
\href{https://doi.org/10.1109/TASL.2009.2025096}{doi:\nolinkurl{10.1109/TASL.2009.2025096}}


\bibitem[Liu et~al\mbox{.}(2017)]%
        {liu2017improved}
\bibfield{author}{\bibinfo{person}{Siqi Liu}, \bibinfo{person}{Zhenhai Zhu},
  \bibinfo{person}{Ning Ye}, \bibinfo{person}{Sergio Guadarrama}, {and}
  \bibinfo{person}{Kevin Murphy}.} \bibinfo{year}{2017}\natexlab{}.
\newblock \showarticletitle{Improved image captioning via policy gradient
  optimization of spider}. In \bibinfo{booktitle}{\emph{Proceedings of the IEEE
  international conference on computer vision}}. \bibinfo{pages}{873--881}.
\newblock


\bibitem[Liu et~al\mbox{.}(2023)]%
        {liu2023geval}
\bibfield{author}{\bibinfo{person}{Yang Liu}, \bibinfo{person}{Dan Iter},
  \bibinfo{person}{Yichong Xu}, \bibinfo{person}{Shuohang Wang},
  \bibinfo{person}{Ruochen Xu}, {and} \bibinfo{person}{Chenguang Zhu}.}
  \bibinfo{year}{2023}\natexlab{}.
\newblock \bibinfo{title}{G-Eval: NLG Evaluation using GPT-4 with Better Human
  Alignment}.
\newblock
\showeprint[arxiv]{2303.16634}~[cs.CL]


\bibitem[Mayer(2009)]%
        {multimedia-mayer}
\bibfield{author}{\bibinfo{person}{Richard~E. Mayer}.}
  \bibinfo{year}{2009}\natexlab{}.
\newblock \bibinfo{booktitle}{\emph{Multimedia Learning}
  (\bibinfo{edition}{2nd} ed.)}.
\newblock \bibinfo{publisher}{Cambridge University Press},
  \bibinfo{address}{USA}.
\newblock
\showISBNx{0521514126}


\bibitem[Ng and Abrecht(2015)]%
        {rouge-1}
\bibfield{author}{\bibinfo{person}{Jun-Ping Ng} {and} \bibinfo{person}{Viktoria
  Abrecht}.} \bibinfo{year}{2015}\natexlab{}.
\newblock \bibinfo{title}{Better Summarization Evaluation with Word Embeddings
  for ROUGE}.
\newblock
\showeprint[arxiv]{1508.06034}~[cs.CL]
\urldef\tempurl%
\url{https://arxiv.org/abs/1508.06034}
\showURL{%
\tempurl}


\bibitem[Nguyen et~al\mbox{.}(2014)]%
        {larochelle}
\bibfield{author}{\bibinfo{person}{Nhu~Van Nguyen}, \bibinfo{person}{Mickal
  Coustaty}, {and} \bibinfo{person}{Jean-Marc Ogier}.}
  \bibinfo{year}{2014}\natexlab{}.
\newblock \showarticletitle{Multi-modal and Cross-Modal for Lecture Videos
  Retrieval}. In \bibinfo{booktitle}{\emph{2014 22nd International Conference
  on Pattern Recognition}}. \bibinfo{pages}{2667--2672}.
\newblock
\href{https://doi.org/10.1109/ICPR.2014.461}{doi:\nolinkurl{10.1109/ICPR.2014.461}}


\bibitem[OpenAI(2023)]%
        {openai2023gpt}
\bibfield{author}{\bibinfo{person}{OpenAI}.} \bibinfo{year}{2023}\natexlab{}.
\newblock \showarticletitle{{GPT}-4 technical report}.
\newblock \bibinfo{journal}{\emph{arXiv preprint arXiv:2303.08774}}
  (\bibinfo{year}{2023}).
\newblock
\urldef\tempurl%
\url{https://arxiv.org/pdf/2303.08774.pdf}
\showURL{%
\tempurl}


\bibitem[OpenAI(2024)]%
        {openai2024gpt4technicalreport}
\bibfield{author}{\bibinfo{person}{OpenAI}.} \bibinfo{year}{2024}\natexlab{}.
\newblock \bibinfo{title}{GPT-4 Technical Report}.
\newblock
\showeprint[arxiv]{2303.08774}~[cs.CL]
\urldef\tempurl%
\url{https://arxiv.org/abs/2303.08774}
\showURL{%
\tempurl}


\bibitem[Owczarzak et~al\mbox{.}(2012)]%
        {rouge-weakness-2}
\bibfield{author}{\bibinfo{person}{Karolina Owczarzak},
  \bibinfo{person}{John~M. Conroy}, \bibinfo{person}{Hoa~Trang Dang}, {and}
  \bibinfo{person}{Ani Nenkova}.} \bibinfo{year}{2012}\natexlab{}.
\newblock \showarticletitle{An Assessment of the Accuracy of Automatic
  Evaluation in Summarization}. In \bibinfo{booktitle}{\emph{Proceedings of
  Workshop on Evaluation Metrics and System Comparison for Automatic
  Summarization}}, \bibfield{editor}{\bibinfo{person}{John~M. Conroy},
  \bibinfo{person}{Hoa~Trang Dang}, \bibinfo{person}{Ani Nenkova}, {and}
  \bibinfo{person}{Karolina Owczarzak}} (Eds.). \bibinfo{publisher}{Association
  for Computational Linguistics}, \bibinfo{address}{Montr{\'e}al, Canada},
  \bibinfo{pages}{1--9}.
\newblock
\urldef\tempurl%
\url{https://aclanthology.org/W12-2601}
\showURL{%
\tempurl}


\bibitem[Panickssery et~al\mbox{.}(2024)]%
        {panickssery2024llm}
\bibfield{author}{\bibinfo{person}{Arjun Panickssery},
  \bibinfo{person}{Samuel~R Bowman}, {and} \bibinfo{person}{Shi Feng}.}
  \bibinfo{year}{2024}\natexlab{}.
\newblock \showarticletitle{Llm evaluators recognize and favor their own
  generations}.
\newblock \bibinfo{journal}{\emph{arXiv preprint arXiv:2404.13076}}
  (\bibinfo{year}{2024}).
\newblock


\bibitem[Papineni et~al\mbox{.}(2002)]%
        {bleu2002}
\bibfield{author}{\bibinfo{person}{Kishore Papineni}, \bibinfo{person}{Salim
  Roukos}, \bibinfo{person}{Todd Ward}, {and} \bibinfo{person}{Wei-Jing Zhu}.}
  \bibinfo{year}{2002}\natexlab{}.
\newblock \showarticletitle{BLEU: a method for automatic evaluation of machine
  translation}. In \bibinfo{booktitle}{\emph{Proceedings of the 40th Annual
  Meeting on Association for Computational Linguistics}} (Philadelphia,
  Pennsylvania) \emph{(\bibinfo{series}{ACL '02})}.
  \bibinfo{publisher}{Association for Computational Linguistics},
  \bibinfo{address}{USA}, \bibinfo{pages}{311–318}.
\newblock
\href{https://doi.org/10.3115/1073083.1073135}{doi:\nolinkurl{10.3115/1073083.1073135}}


\bibitem[Pereira et~al\mbox{.}(2016)]%
        {pereira2016assessment}
\bibfield{author}{\bibinfo{person}{Diana Pereira},
  \bibinfo{person}{Maria~Assun{\c{c}}{\~a}o Flores}, {and}
  \bibinfo{person}{Laila Niklasson}.} \bibinfo{year}{2016}\natexlab{}.
\newblock \showarticletitle{Assessment revisited: a review of research in
  Assessment and Evaluation in Higher Education}.
\newblock \bibinfo{journal}{\emph{Assessment \& Evaluation in Higher
  Education}} \bibinfo{volume}{41}, \bibinfo{number}{7} (\bibinfo{year}{2016}),
  \bibinfo{pages}{1008--1032}.
\newblock


\bibitem[Shen et~al\mbox{.}(2023a)]%
        {shen-etal-2023-large}
\bibfield{author}{\bibinfo{person}{Chenhui Shen}, \bibinfo{person}{Liying
  Cheng}, \bibinfo{person}{Xuan-Phi Nguyen}, \bibinfo{person}{Yang You}, {and}
  \bibinfo{person}{Lidong Bing}.} \bibinfo{year}{2023}\natexlab{a}.
\newblock \showarticletitle{Large Language Models are Not Yet Human-Level
  Evaluators for Abstractive Summarization}. In
  \bibinfo{booktitle}{\emph{Findings of the Association for Computational
  Linguistics: EMNLP 2023}}, \bibfield{editor}{\bibinfo{person}{Houda Bouamor},
  \bibinfo{person}{Juan Pino}, {and} \bibinfo{person}{Kalika Bali}} (Eds.).
  \bibinfo{publisher}{Association for Computational Linguistics},
  \bibinfo{address}{Singapore}, \bibinfo{pages}{4215--4233}.
\newblock
\href{https://doi.org/10.18653/v1/2023.findings-emnlp.278}{doi:\nolinkurl{10.18653/v1/2023.findings-emnlp.278}}


\bibitem[Shen et~al\mbox{.}(2023b)]%
        {shen-etal-2023-human-level}
\bibfield{author}{\bibinfo{person}{Chenhui Shen}, \bibinfo{person}{Liying
  Cheng}, \bibinfo{person}{Xuan-Phi Nguyen}, \bibinfo{person}{Yang You}, {and}
  \bibinfo{person}{Lidong Bing}.} \bibinfo{year}{2023}\natexlab{b}.
\newblock \showarticletitle{Large Language Models are Not Yet Human-Level
  Evaluators for Abstractive Summarization}. In
  \bibinfo{booktitle}{\emph{Findings of the Association for Computational
  Linguistics: EMNLP 2023}}, \bibfield{editor}{\bibinfo{person}{Houda Bouamor},
  \bibinfo{person}{Juan Pino}, {and} \bibinfo{person}{Kalika Bali}} (Eds.).
  \bibinfo{publisher}{Association for Computational Linguistics},
  \bibinfo{address}{Singapore}, \bibinfo{pages}{4215--4233}.
\newblock
\href{https://doi.org/10.18653/v1/2023.findings-emnlp.278}{doi:\nolinkurl{10.18653/v1/2023.findings-emnlp.278}}


\bibitem[Somayeh et~al\mbox{.}(2016)]%
        {somayeh2016effectiveness}
\bibfield{author}{\bibinfo{person}{Mousazadeh Somayeh}, \bibinfo{person}{Maryam
  Dehghani}, \bibinfo{person}{Farzaneh Mozaffari},
  \bibinfo{person}{Seideh~Madineh Ghasemnegad}, \bibinfo{person}{Hamideh
  Hakimi}, {and} \bibinfo{person}{Bagherian Samaneh}.}
  \bibinfo{year}{2016}\natexlab{}.
\newblock \showarticletitle{The effectiveness of E-learning in learning: A
  review of the literature}.
\newblock \bibinfo{journal}{\emph{International journal of medical research \&
  health sciences}} \bibinfo{volume}{5}, \bibinfo{number}{2}
  (\bibinfo{year}{2016}), \bibinfo{pages}{86--91}.
\newblock


\bibitem[Strauss et~al\mbox{.}(2011)]%
        {strauss2011optimizing}
\bibfield{author}{\bibinfo{person}{Judy Strauss}, \bibinfo{person}{Hope
  Corrigan}, {and} \bibinfo{person}{Charles~F Hofacker}.}
  \bibinfo{year}{2011}\natexlab{}.
\newblock \showarticletitle{Optimizing student learning: Examining the use of
  presentation slides}.
\newblock \bibinfo{journal}{\emph{Marketing Education Review}}
  \bibinfo{volume}{21}, \bibinfo{number}{2} (\bibinfo{year}{2011}),
  \bibinfo{pages}{151--162}.
\newblock


\bibitem[Stureborg et~al\mbox{.}(2024)]%
        {stureborg2024large}
\bibfield{author}{\bibinfo{person}{Rickard Stureborg},
  \bibinfo{person}{Dimitris Alikaniotis}, {and} \bibinfo{person}{Yoshi
  Suhara}.} \bibinfo{year}{2024}\natexlab{}.
\newblock \showarticletitle{Large language models are inconsistent and biased
  evaluators}.
\newblock \bibinfo{journal}{\emph{arXiv preprint arXiv:2405.01724}}
  (\bibinfo{year}{2024}).
\newblock


\bibitem[Vasilyev et~al\mbox{.}(2020)]%
        {vasilyev-etal-2020-fill}
\bibfield{author}{\bibinfo{person}{Oleg Vasilyev}, \bibinfo{person}{Vedant
  Dharnidharka}, {and} \bibinfo{person}{John Bohannon}.}
  \bibinfo{year}{2020}\natexlab{}.
\newblock \showarticletitle{Fill in the {BLANC}: Human-free quality estimation
  of document summaries}. In \bibinfo{booktitle}{\emph{Proceedings of the First
  Workshop on Evaluation and Comparison of NLP Systems}},
  \bibfield{editor}{\bibinfo{person}{Steffen Eger}, \bibinfo{person}{Yang Gao},
  \bibinfo{person}{Maxime Peyrard}, \bibinfo{person}{Wei Zhao}, {and}
  \bibinfo{person}{Eduard Hovy}} (Eds.). \bibinfo{publisher}{Association for
  Computational Linguistics}, \bibinfo{address}{Online},
  \bibinfo{pages}{11--20}.
\newblock
\href{https://doi.org/10.18653/v1/2020.eval4nlp-1.2}{doi:\nolinkurl{10.18653/v1/2020.eval4nlp-1.2}}


\bibitem[Vedantam et~al\mbox{.}(2015)]%
        {vedantam2015cider}
\bibfield{author}{\bibinfo{person}{Ramakrishna Vedantam}, \bibinfo{person}{C
  Lawrence~Zitnick}, {and} \bibinfo{person}{Devi Parikh}.}
  \bibinfo{year}{2015}\natexlab{}.
\newblock \showarticletitle{Cider: Consensus-based image description
  evaluation}. In \bibinfo{booktitle}{\emph{Proceedings of the IEEE conference
  on computer vision and pattern recognition}}. \bibinfo{pages}{4566--4575}.
\newblock


\bibitem[Wang and Chan(2019)]%
        {wang2019describing}
\bibfield{author}{\bibinfo{person}{Qingzhong Wang} {and}
  \bibinfo{person}{Antoni~B Chan}.} \bibinfo{year}{2019}\natexlab{}.
\newblock \showarticletitle{Describing like humans: on diversity in image
  captioning}. In \bibinfo{booktitle}{\emph{Proceedings of the IEEE/CVF
  Conference on Computer Vision and Pattern Recognition}}.
  \bibinfo{pages}{4195--4203}.
\newblock


\bibitem[Wang and Demszky(2023)]%
        {wang-demszky-2023-chatgpt}
\bibfield{author}{\bibinfo{person}{Rose Wang} {and} \bibinfo{person}{Dorottya
  Demszky}.} \bibinfo{year}{2023}\natexlab{}.
\newblock \showarticletitle{Is {C}hat{GPT} a Good Teacher Coach? Measuring
  Zero-Shot Performance For Scoring and Providing Actionable Insights on
  Classroom Instruction}. In \bibinfo{booktitle}{\emph{Proceedings of the 18th
  Workshop on Innovative Use of NLP for Building Educational Applications (BEA
  2023)}}, \bibfield{editor}{\bibinfo{person}{Ekaterina Kochmar},
  \bibinfo{person}{Jill Burstein}, \bibinfo{person}{Andrea Horbach},
  \bibinfo{person}{Ronja Laarmann-Quante}, \bibinfo{person}{Nitin Madnani},
  \bibinfo{person}{Ana{\"\i}s Tack}, \bibinfo{person}{Victoria Yaneva},
  \bibinfo{person}{Zheng Yuan}, {and} \bibinfo{person}{Torsten Zesch}} (Eds.).
  \bibinfo{publisher}{Association for Computational Linguistics},
  \bibinfo{address}{Toronto, Canada}, \bibinfo{pages}{626--667}.
\newblock
\href{https://doi.org/10.18653/v1/2023.bea-1.53}{doi:\nolinkurl{10.18653/v1/2023.bea-1.53}}


\bibitem[Wang et~al\mbox{.}(2024)]%
        {wang2024pandalm}
\bibfield{author}{\bibinfo{person}{Yidong Wang}, \bibinfo{person}{Zhuohao Yu},
  \bibinfo{person}{Zhengran Zeng}, \bibinfo{person}{Linyi Yang},
  \bibinfo{person}{Cunxiang Wang}, \bibinfo{person}{Hao Chen},
  \bibinfo{person}{Chaoya Jiang}, \bibinfo{person}{Rui Xie},
  \bibinfo{person}{Jindong Wang}, \bibinfo{person}{Xing Xie},
  \bibinfo{person}{Wei Ye}, \bibinfo{person}{Shikun Zhang}, {and}
  \bibinfo{person}{Yue Zhang}.} \bibinfo{year}{2024}\natexlab{}.
\newblock \bibinfo{title}{PandaLM: An Automatic Evaluation Benchmark for LLM
  Instruction Tuning Optimization}.
\newblock
\showeprint[arxiv]{2306.05087}~[cs.CL]


\bibitem[Wei et~al\mbox{.}(2022)]%
        {wei2022chain}
\bibfield{author}{\bibinfo{person}{Jason Wei}, \bibinfo{person}{Xuezhi Wang},
  \bibinfo{person}{Dale Schuurmans}, \bibinfo{person}{Maarten Bosma},
  \bibinfo{person}{Fei Xia}, \bibinfo{person}{Ed Chi}, \bibinfo{person}{Quoc~V
  Le}, \bibinfo{person}{Denny Zhou}, {et~al\mbox{.}}}
  \bibinfo{year}{2022}\natexlab{}.
\newblock \showarticletitle{Chain-of-thought prompting elicits reasoning in
  large language models}.
\newblock \bibinfo{journal}{\emph{Advances in neural information processing
  systems}}  \bibinfo{volume}{35} (\bibinfo{year}{2022}),
  \bibinfo{pages}{24824--24837}.
\newblock


\bibitem[Xie et~al\mbox{.}(2024)]%
        {xie-etal-2024-doclens}
\bibfield{author}{\bibinfo{person}{Yiqing Xie}, \bibinfo{person}{Sheng Zhang},
  \bibinfo{person}{Hao Cheng}, \bibinfo{person}{Pengfei Liu},
  \bibinfo{person}{Zelalem Gero}, \bibinfo{person}{Cliff Wong},
  \bibinfo{person}{Tristan Naumann}, \bibinfo{person}{Hoifung Poon}, {and}
  \bibinfo{person}{Carolyn Rose}.} \bibinfo{year}{2024}\natexlab{}.
\newblock \showarticletitle{{D}oc{L}ens: Multi-aspect Fine-grained Medical Text
  Evaluation}. In \bibinfo{booktitle}{\emph{Proceedings of the 62nd Annual
  Meeting of the Association for Computational Linguistics (Volume 1: Long
  Papers)}}, \bibfield{editor}{\bibinfo{person}{Lun-Wei Ku},
  \bibinfo{person}{Andre Martins}, {and} \bibinfo{person}{Vivek Srikumar}}
  (Eds.). \bibinfo{publisher}{Association for Computational Linguistics},
  \bibinfo{address}{Bangkok, Thailand}, \bibinfo{pages}{649--679}.
\newblock
\urldef\tempurl%
\url{https://aclanthology.org/2024.acl-long.39}
\showURL{%
\tempurl}


\bibitem[Xie et~al\mbox{.}(2023)]%
        {xie-etal-2023-deltascore}
\bibfield{author}{\bibinfo{person}{Zhuohan Xie}, \bibinfo{person}{Miao Li},
  \bibinfo{person}{Trevor Cohn}, {and} \bibinfo{person}{Jey Lau}.}
  \bibinfo{year}{2023}\natexlab{}.
\newblock \showarticletitle{{D}elta{S}core: Fine-Grained Story Evaluation with
  Perturbations}. In \bibinfo{booktitle}{\emph{Findings of the Association for
  Computational Linguistics: EMNLP 2023}},
  \bibfield{editor}{\bibinfo{person}{Houda Bouamor}, \bibinfo{person}{Juan
  Pino}, {and} \bibinfo{person}{Kalika Bali}} (Eds.).
  \bibinfo{publisher}{Association for Computational Linguistics},
  \bibinfo{address}{Singapore}, \bibinfo{pages}{5317--5331}.
\newblock
\href{https://doi.org/10.18653/v1/2023.findings-emnlp.353}{doi:\nolinkurl{10.18653/v1/2023.findings-emnlp.353}}


\bibitem[Xu et~al\mbox{.}(2023)]%
        {xu-etal-2023-instructscore}
\bibfield{author}{\bibinfo{person}{Wenda Xu}, \bibinfo{person}{Danqing Wang},
  \bibinfo{person}{Liangming Pan}, \bibinfo{person}{Zhenqiao Song},
  \bibinfo{person}{Markus Freitag}, \bibinfo{person}{William Wang}, {and}
  \bibinfo{person}{Lei Li}.} \bibinfo{year}{2023}\natexlab{}.
\newblock \showarticletitle{{INSTRUCTSCORE}: Towards Explainable Text
  Generation Evaluation with Automatic Feedback}. In
  \bibinfo{booktitle}{\emph{Proceedings of the 2023 Conference on Empirical
  Methods in Natural Language Processing}},
  \bibfield{editor}{\bibinfo{person}{Houda Bouamor}, \bibinfo{person}{Juan
  Pino}, {and} \bibinfo{person}{Kalika Bali}} (Eds.).
  \bibinfo{publisher}{Association for Computational Linguistics},
  \bibinfo{address}{Singapore}, \bibinfo{pages}{5967--5994}.
\newblock
\href{https://doi.org/10.18653/v1/2023.emnlp-main.365}{doi:\nolinkurl{10.18653/v1/2023.emnlp-main.365}}


\bibitem[Yao et~al\mbox{.}(2024)]%
        {yao2024minicpm}
\bibfield{author}{\bibinfo{person}{Yuan Yao}, \bibinfo{person}{Tianyu Yu},
  \bibinfo{person}{Ao Zhang}, \bibinfo{person}{Chongyi Wang},
  \bibinfo{person}{Junbo Cui}, \bibinfo{person}{Hongji Zhu},
  \bibinfo{person}{Tianchi Cai}, \bibinfo{person}{Haoyu Li},
  \bibinfo{person}{Weilin Zhao}, \bibinfo{person}{Zhihui He}, {et~al\mbox{.}}}
  \bibinfo{year}{2024}\natexlab{}.
\newblock \showarticletitle{MiniCPM-V: A GPT-4V Level MLLM on Your Phone}.
\newblock \bibinfo{journal}{\emph{arXiv preprint arXiv:2408.01800}}
  (\bibinfo{year}{2024}).
\newblock


\bibitem[Yi et~al\mbox{.}(2020)]%
        {yi-etal-2020-improving}
\bibfield{author}{\bibinfo{person}{Yanzhi Yi}, \bibinfo{person}{Hangyu Deng},
  {and} \bibinfo{person}{Jinglu Hu}.} \bibinfo{year}{2020}\natexlab{}.
\newblock \showarticletitle{Improving Image Captioning Evaluation by
  Considering Inter References Variance}. In
  \bibinfo{booktitle}{\emph{Proceedings of the 58th Annual Meeting of the
  Association for Computational Linguistics}},
  \bibfield{editor}{\bibinfo{person}{Dan Jurafsky}, \bibinfo{person}{Joyce
  Chai}, \bibinfo{person}{Natalie Schluter}, {and} \bibinfo{person}{Joel
  Tetreault}} (Eds.). \bibinfo{publisher}{Association for Computational
  Linguistics}, \bibinfo{address}{Online}, \bibinfo{pages}{985--994}.
\newblock
\href{https://doi.org/10.18653/v1/2020.acl-main.93}{doi:\nolinkurl{10.18653/v1/2020.acl-main.93}}


\bibitem[Yu et~al\mbox{.}(2021)]%
        {yu2021mooccubex}
\bibfield{author}{\bibinfo{person}{Jifan Yu}, \bibinfo{person}{Yuquan Wang},
  \bibinfo{person}{Qingyang Zhong}, \bibinfo{person}{Gan Luo},
  \bibinfo{person}{Yiming Mao}, \bibinfo{person}{Kai Sun},
  \bibinfo{person}{Wenzheng Feng}, \bibinfo{person}{Wei Xu},
  \bibinfo{person}{Shulin Cao}, \bibinfo{person}{Kaisheng Zeng},
  {et~al\mbox{.}}} \bibinfo{year}{2021}\natexlab{}.
\newblock \showarticletitle{MOOCCubeX: a large knowledge-centered repository
  for adaptive learning in MOOCs}. In \bibinfo{booktitle}{\emph{Proceedings of
  the 30th ACM International Conference on Information \& Knowledge
  Management}}. \bibinfo{pages}{4643--4652}.
\newblock


\bibitem[Yuan et~al\mbox{.}(2023)]%
        {yuan2023batcheval}
\bibfield{author}{\bibinfo{person}{Peiwen Yuan}, \bibinfo{person}{Shaoxiong
  Feng}, \bibinfo{person}{Yiwei Li}, \bibinfo{person}{Xinglin Wang},
  \bibinfo{person}{Boyuan Pan}, \bibinfo{person}{Heda Wang}, {and}
  \bibinfo{person}{Kan Li}.} \bibinfo{year}{2023}\natexlab{}.
\newblock \bibinfo{title}{BatchEval: Towards Human-like Text Evaluation}.
\newblock
\showeprint[arxiv]{2401.00437}~[cs.CL]
\urldef\tempurl%
\url{https://arxiv.org/abs/2401.00437}
\showURL{%
\tempurl}


\bibitem[Zhang et~al\mbox{.}(2024b)]%
        {rouge-3}
\bibfield{author}{\bibinfo{person}{Ming Zhang}, \bibinfo{person}{Chengzhang
  Li}, \bibinfo{person}{Meilin Wan}, \bibinfo{person}{Xuejun Zhang}, {and}
  \bibinfo{person}{Qingwei Zhao}.} \bibinfo{year}{2024}\natexlab{b}.
\newblock \showarticletitle{ROUGE-SEM: Better evaluation of summarization using
  ROUGE combined with semantics}.
\newblock \bibinfo{journal}{\emph{Expert Systems with Applications}}
  \bibinfo{volume}{237} (\bibinfo{year}{2024}), \bibinfo{pages}{121364}.
\newblock
\showISSN{0957-4174}
\href{https://doi.org/10.1016/j.eswa.2023.121364}{doi:\nolinkurl{10.1016/j.eswa.2023.121364}}


\bibitem[Zhang et~al\mbox{.}(2024c)]%
        {zhang2024rouge}
\bibfield{author}{\bibinfo{person}{Ming Zhang}, \bibinfo{person}{Chengzhang
  Li}, \bibinfo{person}{Meilin Wan}, \bibinfo{person}{Xuejun Zhang}, {and}
  \bibinfo{person}{Qingwei Zhao}.} \bibinfo{year}{2024}\natexlab{c}.
\newblock \showarticletitle{ROUGE-SEM: Better evaluation of summarization using
  ROUGE combined with semantics}.
\newblock \bibinfo{journal}{\emph{Expert Systems with Applications}}
  \bibinfo{volume}{237} (\bibinfo{year}{2024}), \bibinfo{pages}{121364}.
\newblock


\bibitem[Zhang et~al\mbox{.}(2020)]%
        {zhang2020bertscore}
\bibfield{author}{\bibinfo{person}{Tianyi Zhang}, \bibinfo{person}{Varsha
  Kishore}, \bibinfo{person}{Felix Wu}, \bibinfo{person}{Kilian~Q. Weinberger},
  {and} \bibinfo{person}{Yoav Artzi}.} \bibinfo{year}{2020}\natexlab{}.
\newblock \bibinfo{title}{BERTScore: Evaluating Text Generation with BERT}.
\newblock
\showeprint[arxiv]{1904.09675}~[cs.CL]


\bibitem[Zhang et~al\mbox{.}(2024a)]%
        {zhang2024towards}
\bibfield{author}{\bibinfo{person}{Weijia Zhang}, \bibinfo{person}{Mohammad
  Aliannejadi}, \bibinfo{person}{Yifei Yuan}, \bibinfo{person}{Jiahuan Pei},
  \bibinfo{person}{Jia-Hong Huang}, {and} \bibinfo{person}{Evangelos
  Kanoulas}.} \bibinfo{year}{2024}\natexlab{a}.
\newblock \showarticletitle{Towards fine-grained citation evaluation in
  generated text: A comparative analysis of faithfulness metrics}.
\newblock \bibinfo{journal}{\emph{arXiv preprint arXiv:2406.15264}}
  (\bibinfo{year}{2024}).
\newblock


\end{thebibliography}

\end{document}